\documentclass{article}

\usepackage{arxiv}

\usepackage[utf8]{inputenc} 
\usepackage[T1]{fontenc}    
\usepackage{hyperref}       
\usepackage{url}            
\usepackage{booktabs}       
\usepackage{amsfonts}       
\usepackage{nicefrac}       
\usepackage{microtype}      
\usepackage{lipsum}
\usepackage{graphicx}
\graphicspath{ {./images/} }

\title{Projection: A Mixed-Initiative Research Process}

\author{
 Austin Silveria \\
  California Polytechnic State University, San Luis Obispo\\
  San Luis Obispo, CA 93405 \\
  \texttt{silveria@calpoly.edu}\\
}

\begin{document}
\maketitle

\begin{abstract}
Communication of dense information between humans and machines is relatively low bandwidth. Many modern search and recommender systems operate as machine learning black boxes, giving little insight as to how they represent information or why they take certain actions. We present Projection, a mixed-initiative interface that aims to increase the bandwidth of communication between humans and machines throughout the research process. The interface supports adding context to searches and visualizing information in multiple dimensions with techniques such as hierarchical clustering and spatial projections. Potential customers have shown interest in the application integrating their research outlining and search processes, enabling them to structure their searches in hierarchies, and helping them visualize related spaces of knowledge.
\end{abstract}

\section{Introduction}
We consider conceptualizing the online research process as a specific instance of the scientific method: posing a question, collecting information to answer the question, organizing the information, and communicating the answer, where the steps are cycled multiple times in the process.

In interviews conducted by the author, we talked to five college students about their research process.
We learned that they begin the research process with an outline of how they think about a topic. They then search Google with targeted queries to find information to support this structure. Along the way, their outline evolves and prompts new searches.
We learned that people are frustrated by the lack of integration between their outline and searches. \emph{I’ve organized my ideas in this outline, but how do I communicate that to Google? How can I search for new information in the context of research I’ve already done?}
We learned that people find it difficult to survey a topic at a high level. \emph{I have this idea of the field that I’m trying to explore, but how can I map that field into subtopics to understand what makes it up? I find a number of articles through my searches, but it’s overwhelming to manually read and organize all of them.}

As an extension to our previous work \cite{DBLP:journals/corr/abs-2007-11198}, we built Projection, a mixed-initiative interface to the research process. The tool clusters documents on a map based on their similarity (as seen in Figure 1), the goal being that similar documents show up on the map next to each other and can be grouped into hierarchical clusters \cite{procluster}. Projection also includes a function to search on the map (as seen in Figure 2), project results in nearby space, and group them into hierarchical clusters \cite{prosearch}.
We demoed the functionality to the same five interviewees to spur discussion about product development.
We asked them: \emph{given this organized knowledge space, how would you want to navigate it? How would you want to organize it yourself? What kind of information would you map out? How is this type of search different from current solutions?}
We learned that they would want to start typing on a map, visualize related topics, and continue searching through those topics, building a hierarchy of what they are thinking and continuing to search in the context of that hierarchy.
They mentioned a side tab with a traditional outline. One could click on the outline to go to that place on the map. One could use the outline to maintain a context of where they are in the knowledge space.
We talked about multiple people interacting with the same information in Projection. \emph{How do other people structure these topics I’m interested in? In the context of a large organization, who should I talk to about this topic? Who has spent a lot of time structuring this realm of information?}
We learned that they would want to explore types of information such as wikipedia, research articles, web articles, and personal documents.
We found that Projection can be compared to traditional search in the following way: instead of a list of pages, Projection visualizes a knowledge space. Instead of searching with a single natural language query, Projection searches with a hierarchy of natural language. Instead of each search being independent, Projection searches come together into one navigable structure.

\begin{figure*}[h]
  \centering
  \includegraphics[width=\linewidth]{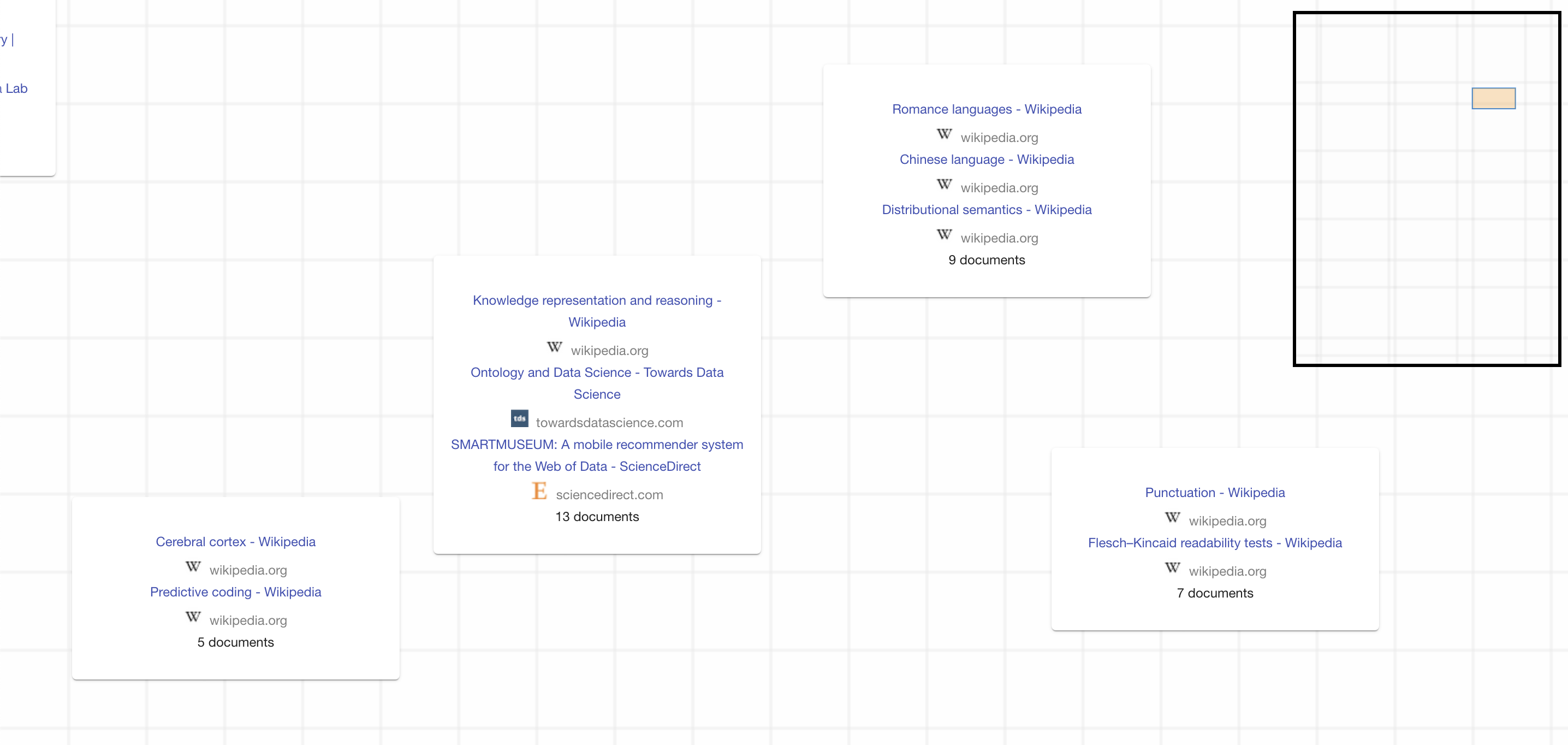}
  \caption{Projection Spatial Document Clustering \cite{procluster}}
  \label{skipthought}
\end{figure*}

\section{Backgound}
\subsection{Data Model}
Consider an entity graph, where the meaning of an entity depends on its relationship to and the meaning of entities connected to it. We apply this data model to Projection such that each piece of information is an entity that can be structured and searched for in the context of other entities. As an example, consider a hierarchy of entities mapping the sub topics of a field. The parent-child connections of the hierarchy communicate the relationship between entities and can be used to find more entities that fit into the structure.

\subsection{Mixed-Initiative Interface}
Mixed-initiative interaction refers to a flexible interaction strategy in which each agent (human or computer) contributes what it is best suited at the most appropriate time \cite{allen_guinn_horvtz_1999}. In applying this to Projection, humans manually structure entities and the computer responds with contextually relevant entities projected into the structure. For example, after creating an entity with the language "investing", the machine might project clusters of entities about types of investments, famous investors, or investment strategies. As more structure is given to the entities by the human, the machine has more context to suggest relevant types of entities.

\subsection{Software Libraries}
React \cite{react} is a JavaScript library Projection uses for building declarative and component-based interfaces. React automatically re renders components when their inputs change, saving developers from manually triggering updates.

Redux \cite{redux} is a predictable state container for JavaScript applications. State is interacted with via a global store exposed by Redux. React components select subsets of this state to render and emit actions to update state based on user interaction. When a component's selected state changes, it is automatically re rendered. With this functionality, Redux serves as a powerful communication model between many declarative functions that need access to the same state.

Mapbox \cite{mapbox} is a JavaScript library for building map applications in the browser. It exposes a powerful map API that developers can interface with to render components in 2D/3D space--the library then manages the entire map once it is rendered: zoom, drag, re renders, etc.

Projection uses AWS Amplify \cite{amplify} to manage web authentication and database hosting. With a single GraphQL schema, Amplify can deploy a GraphQL server and multiple DynamoDB tables to fully support queries and mutations in a managed cloud environment.

On the backend we use Tensorflow's Universal Sentence Encoder \cite{use} to compute language embeddings and Facebook's FAISS \cite{faiss} for large scale vector operations such as similarity search.

\section{Related Work}

\subsection{Data Model}
The entity graph data model can be seen in various environments of computationally defining meaning: the meaning of a word is the meaning of words around it \cite{mikolov2013efficient}, the attention computation is a fully connected, directed graph with learned connection weights \cite{DBLP:journals/corr/VaswaniSPUJGKP17}, the meaning of an Instagram account is the accounts it interacts with \cite{instagram-recommender}, and the meaning of a YouTube video is the context in which it was watched \cite{45530}.

People also consciously organize meaning with discrete connections. A hierarchical research outline is a tree of natural language, document citations form a document graph, and digital friends lists connect us to others online. Unconsciously, a perspective could be that our brains define meaning by relating information in multiple dimensions. Visual perception relates points of space based on position and color. Auditory perception detects changes in pressure of the environment across time. Our brain relates information across all of these dimensions unconsciously, and we use this high dimensional perception to act.

Work done by Yang et al. \cite{DBLP:journals/corr/abs-1806-05662} explores the value of decoupling the learning of entity relationships and semantic meaning. Compared to self-attention, which fuses the computation of entity structure and meaning into one network, their work decouples the two, allowing the learned entity structure to transfer to new tasks where different features may be needed to model semantic meaning. This is interesting from our perspective since Projection aims to be a mixed-initiative interface that allows people to structure information in any format. Structuring information in different ways or structuring different types of information may be analogous to different tasks, warranting consideration of different semantic features. Structures created by everyone using Projection can also be valuable training data for computing entity meaning.

\subsection{Mixed-Initiative Interface}
Examples of mixed-initiative interfaces include recommender systems of popular internet applications. People interact with social connections online and Instagram recommends new social connections \cite{instagram-recommender}. People watch YouTube videos, and YouTube learns from this context to suggest new videos \cite{45530}. People search Google, and the model learns from this history to auto complete relevant queries.

\begin{figure*}[h]
  \centering
  \includegraphics[width=\linewidth]{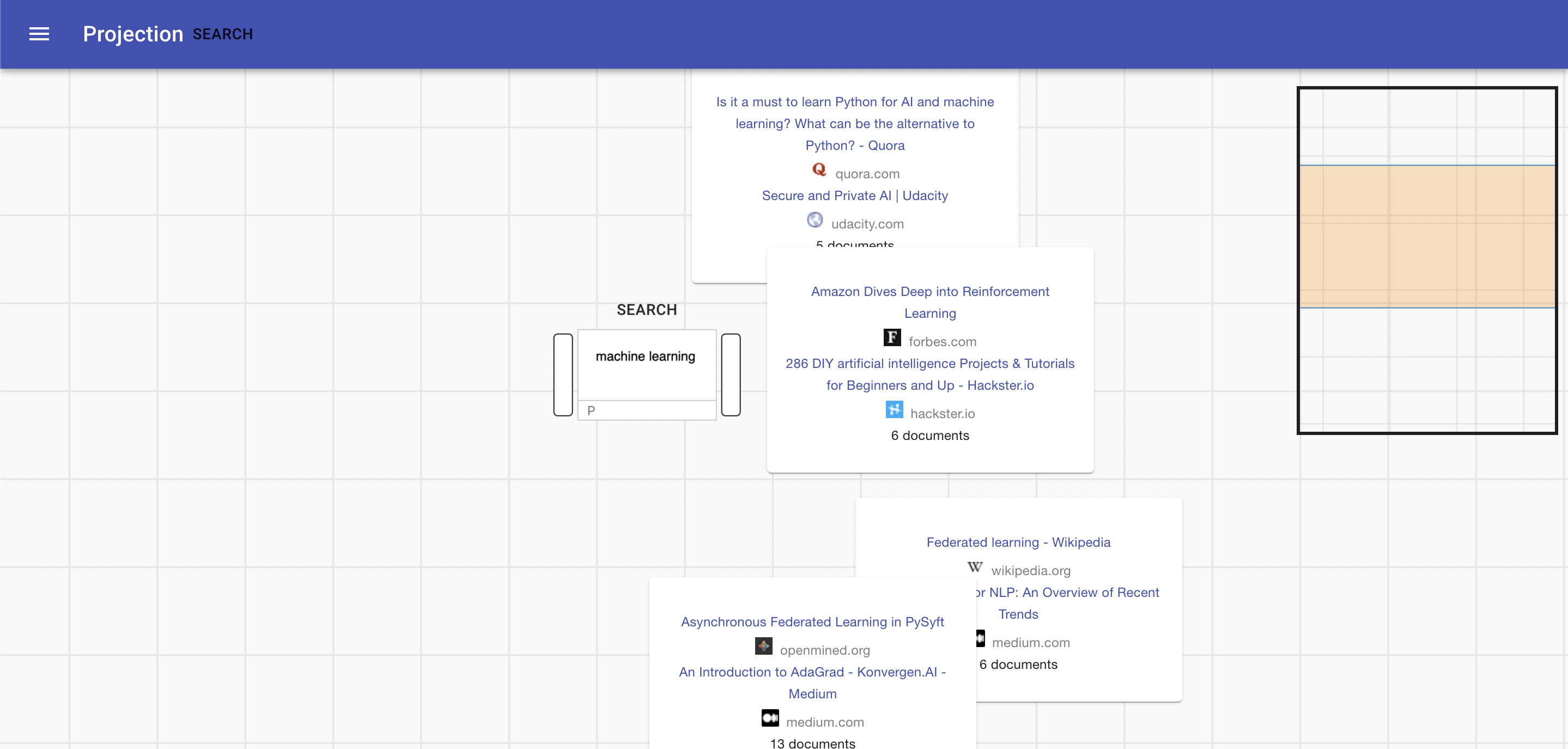}
  \caption{Projection Search \cite{prosearch}}
  \label{skipthought}
\end{figure*}

Many existing mixed initiative interfaces are imbalanced--the machine learning model operates in a higher dimension than the human guiding it, leading to a lower bandwidth of communication between the model and human. It is difficult to understand the model’s interpretation of meaning and guide its suggestion making.

To increase the bandwidth of communication between the model and human, a mixed-initiative interface can be designed with multiple dimensions: searching based on a hierarchy of language, clustering search results to communicate topics, projecting entities on a 2D/3D space to communicate their relationships, or at the highest dimension with a brain-machine interface. Even a small increase in dimension of the human side of the interface can exponentially increase the bandwidth of communication, improving the shared meaning representation between model and human.

\section{System}

We used AWS Amplify \cite{amplify} to generate a set of DynamoDB tables to track object state and a GraphQL server to support object queries and mutations. We define a single GraphQL schema with our application's data model and leverage Amplify to manage the rest.

Outside of Amplify, we split the Projection architecture into two components: a web interface to search, explore, and structure information and a backend to support stateful computation of data such as document similarities, clusters, and spatial entity projections. Our full architecture can be seen in Figure 3.

\subsection{Interface}
The interface supports two core actions: humans creating/structuring entities and the model suggesting new entities/clusters in the context of that structure. People drive the structuring of information and the model responds with information that meshes with that structure.

To track application state, we define the data model shown in Table 1. We use React \cite{react}, Redux \cite{redux}, and Mapbox \cite{mapbox} to build modular interface components that react independently to application state changes and mirror the data model. The \emph{Menu} supports creating and selecting maps. The \emph{Map} component reacts to a map being selected in the \emph{Menu} component, queries the map's entities, and renders multiple \emph{Entity} components on the map. The \emph{Entity} and \emph{Document} components query and render model data. Additional \emph{Cluster} components render sets of \emph{Entity}s or \emph{Document}s as hierarchical clusters.

This modular component architecture on top of React and Redux allow us to easily test new components. A new component can be inserted anywhere in the interface model, Redux can be leveraged to query state of other components and communicate state updates, and we can let React handle efficient re renders based on state changes.

\begin{table*}
  \caption{Projection Data Model}
  \label{tab:search}
  \centering
    \begin{tabular}{ |p{3cm}|p{3cm}|p{3cm}|p{3cm}|  }
     \hline
     Menu & Map & Entity & Document\\
     \hline
     selectedMapId & mapId & entityId & documentId\\
      & name  & mapId & title\\
      & entityIds & parentEntityId & url\\
         & & childrenEntityIds & text\\
      & & coordinates &\\
      & & text &\\
     \hline
    \end{tabular}
\end{table*}

\subsection{Backend}

At the core of Projection's backend is FAISS \cite{faiss}, a library used to efficiently store dense vectors and compute vector similarities. A web server sits on top of FAISS and exposes \emph{add} and \emph{query} operations.

We organize Projection's backend computation into a functional pipeline for each operation. \emph{add} supports loading new language entities into FAISS by first computing a vector representation of the language with the Universal Sentence Encoder \cite{use}, then loading it into the index. The \emph{query} operation receives an entity tree as input and executes the following pipeline to construct a response: embed the text of each entity, search for documents/entities in context of the tree, cluster the resulting documents/entities, and project the response entities in space near the input entity tree.

\begin{figure*}[h]
  \centering
  \includegraphics[width=\linewidth]{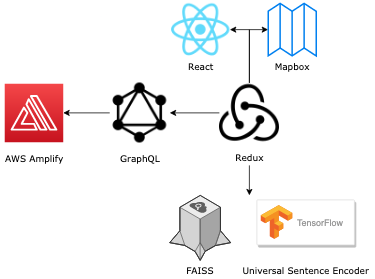}
  \caption{Projection Architecture}
  \label{skipthought}
\end{figure*}

One strategy we consider to search in hierarchical context is as follows: find neighbors of the root entity, find neighbors of children within the parent level's neighborhood space, and continue in a depth first search fashion. This would allow human operators to search by "partitioning" the meaning space with their structured entity hierarchy. With the functional pipeline architecture, new strategies can easily be injected into the pipeline for testing.

To cluster entities represented by dense vectors in 2D space we first reduce the dimensionality of the vectors from 512 to 2 with T-SNE, then apply agglomerative clustering to find hiearchical structure.

Our goal for projecting entities is to move a set of entities in 2D space and scale their distances while maintaining directional relationships. Our initial multiplication transform is acceptable for testing, but future work should use a known mathematical operation for maximum precision.

\subsection{MVP}
To create an initial MVP based on the architecture described above, we gathered web pages from our personal search history and loaded them into the backend. Web pages can be explored with hierarchical clusters on the map \cite{procluster} and searched for by typing on the map \cite{prosearch}.

\section{Evaluation}

To evaluate our product, we demoed the functionality to five college students as discussed in the introduction. All interviewees were excited by the potential of the product, but had a different purpose in mind than a traditional search engine. They would use Projection when they didn't have a specific answer in mind. They would want start typing on the map and explore the different perspectives of their natural language query. They would connect their initial query to new information, and continue their search/exploration of the knowledge space.

For Projection to be usable, the interviewees highlighted a few points of important functionality. First, it needs to contain the data that they are interested in exploring. For those we interviewed, this was Wikipedia entries, research articles, web articles, and personal documents. Filters should be available to narrow the information considered to a certain type. They need to be able to maintain context of where they are in the knowledge space. It was easy to get lost on the map because parent clusters split into child clusters on zoom, and weren't visible for reference after. They mentioned being able to see multiple levels of parent-child clusters on the map at the same time in order to maintain depth context. They also mentioned a sidebar with a traditional bullet point outline. This outline could represent their mapped entity structure in a familiar format and highlight where they currently are in the knowledge space. The outline would be interactive: clicks take one to that location on the map, and the current location on the map is highlighted in the outline. Another piece of requested functionality was group search. \emph{I have this group of entities here and I want to find more entities that fit in the group}. In the future, this can easily be implemented with similarity search and tested in our backend's query pipeline.

\section{Conclusion}

We present Projection, a mixed-initiative interface to the research process. Through hierarchical clustering, spatial projections, and an intuitive knowledge structure, Projection can increase the bandwidth of communication between humans and machines, a central issue in today's interfaces to machine learning. The human and machine represent information as an entity graph, which can be used to communicate structure and meaning by both parties. Our scalable software architecture will allow rapid experimentation of new features and parallel development going forward.

\section{Future Work}

We have a strong roadmap of future work ahead based on our interview discussions. Likely the most important step going forward is expanding the types and scale of information accessible in Projection beyond our personal search history. Implementing a web scraper to collect web pages can automate this process, but customers should also be able to import their own documents or trigger a scrape of a website. This would allow potential customers to test the tool with information that is valuable to them, making their experience and feedback more authentic.

Experimenting with the interface is another crucial step going forward. We will need to discover the right amount of information to display at one time, how to communicate where one is in their defined structure and the overall knowledge space, and how to communicate the contents of varying types of entity clusters. To conduct real tests, this should be done after additional information is loaded into Projection so that potential customers can use the product for practical purposes.

With the ability to conduct real tests, we will also be able experiment with various machine learning methods on the backend. Testing different search strategies and machine learning models will help us identify what solves customer problems the best.

We will also continue abstracting our architecture with the goal of enabling the lowest cost of experimentation. The cost of experimentation can be broken into three parts: complexity of implementation, effort required to implement, and time required to implement. A set of functions that react independently to global state changes has proven to be a powerful architecture, one that we are looking to adopt across our entire software stack. Functions in the frontend or backend, across any number of physical machines, communicate via access-controlled namespaces of a global state store and react to queried state changes to produce additional state changes.

\section{Acknowledgements}

We would like to thank Foaad Khosmood for advising our research.

\bibliographystyle{unsrt}
\bibliography{explore}

\end{document}